\documentclass[conference]{IEEEtran}
\IEEEoverridecommandlockouts
\usepackage{placeins}
\usepackage{amsmath,amssymb,amsfonts}
\usepackage{algorithmic}
\usepackage{graphicx}
\usepackage{textcomp}
\usepackage{xcolor}
\usepackage{easy-todo}
\usepackage{float}
\usepackage{tikz}
\usepackage{subfig}
\usepackage[style=ieee]{biblatex}
\bibliography{include/backmatter/refs.bib}

\begin{document}

\title{Extended Object Tracking Using Sets Of Trajectories with a PHD Filter\\
}

\author{\IEEEauthorblockN{Jakob Sjudin, Martin Marcusson}
\IEEEauthorblockA{\textit{SafeRadar Research Sweden} \\
Gothenburg, Sweden \\
firstname.lastname@saferadar.se}
\and
\IEEEauthorblockN{Lennart Svensson, Lars Hammarstrand}
\IEEEauthorblockA{\textit{Dept. of Electrical Engineering} \\
\textit{Chalmers University of Technology}\\
Gothenburg, Sweden \\
firstname.lastname@chalmers.se}
}

\maketitle


\begin{abstract}
PHD filtering is a common and effective multiple object tracking (MOT) algorithm used in scenarios where the number of objects and their states are unknown. 
In scenarios where each object can generate multiple measurements per scan, some PHD filters can estimate the extent of the objects as well as their kinematic properties. 
Most of these approaches are, however, not able to inherently estimate trajectories and rely on ad-hoc methods, such as different labeling schemes, to build trajectories from the state estimates. 
This paper presents a Gamma Gaussian inverse Wishart mixture PHD filter that can directly estimate sets of trajectories of extended targets by expanding previous research on tracking sets of trajectories for point source objects to handle extended objects. 
The new filter is compared to an existing extended PHD filter that uses a labeling scheme to build trajectories, and it is shown that the new filter can estimate object trajectories more reliably.

\end{abstract}

\begin{IEEEkeywords}
Multiple object tracking, Extended objects, Gamma Gaussian inverse Wishart, Trajectories, Bayesian smoothing, Random finite sets, PHD filtering.
\end{IEEEkeywords}

\section{Introduction} \label{ch:intro}


Tracking of multiple moving objects is important in many different areas, e.g., surveillance of aircraft or self-driving vehicles. Tracking in this context refers to estimating the number -- and the kinematic properties (position, velocity and heading) -- of the objects currently observed by one or more sensors, such as radars, lidars and cameras. A key challenge is that the measurements obtained from these sensors are unlabeled, corrupted by noise and clutter, and suffers from missed detections. The aim of a multiple object tracking (MOT) algorithm is, thus, to correctly associate observations to the unknown and time-varying number of objects from which their kinematic state can be estimated. This is a challenging problem for which many different methods have been proposed in the literature, e.g., Multiple Hypothesis Tracking (MHT) \cite{blackman2004multiple}, Joint Probabilistic Data Association (JPDA) \cite{5338565}, Probabilistic Multiple Hypothesis Tracking (PMHT) \cite{streit1995probabilistic} and Random Finite Set (RFS) methods \cite{Mahler2014}. 

In recent years, much focus has been on the RFS-based methods that treat the tracking problem in the theoretical framework of random finite sets \cite{Mahler2007statistical}. Among these, the probability hypothesis density (PHD) filter is among the more computationally efficient and have been proven to give good results in many different applications \cite{lundquist2010road, garcia2018phd, Granstrom2015}. The PHD filter describes the multi-object posterior using the first moment of the RFS and specific implementations are derived under several different modeling assumptions. For example, the original Gaussian mixture PHD (GMPHD) filter \cite{GMPHD-filter} that assumes point-source objects (at most one measurement from each object). The point source assumption is relaxed in the Gaussian inverse Wishart PHD (GIWPHD) filter \cite{Lundquist2013} and the Gamma Gaussian inverse Wishart PHD (GGIWPHD) filter \cite{Granstrom2015}. These filters are designed to handle extended objects (each object can generate multiple measurements) that are common in modern high resolution sensors. In comparison to point-source objects, this allows for direct estimation of the spatial extent as well as the orientation of the object. 

In many applications, one is not only interested in knowing the current position of an object, but also where an object has been in the past. One such example is in supervised learning where large datasets with time series of object detections need to be annotated with ground truth information. By directly extracting (estimating) object trajectories from the time series data the annotation process could be sped up significantly by, e.g., only annotating every other frame/scan.

The classical formulation of PHD filters can not inherently build trajectories. There are, however, different ad-hoc methods to label object states and adding unique tags to the PHD components \cite{4086095,7338659,5259179}. Although these modifications make it possible to build trajectories, it is an unreliable method which can lead to unstable behavior with track switching, false targets, and missed detections. There are, however, more recent RFS-based methods that are shown to be less prone to these behaviors \cite{Garcia-Fernandez2016,Garcia-Fernandez2018}. These methods use a more formal and direct way of estimating object trajectories by formulating the multi-object state as a set of object trajectories. 
Examples of such methods are the trajectory PHD (TPHD) filter \cite{Garcia-Fernandez2018} for point-source objects and the trajectory Poisson multi-Bernoulli mixture filter for extended objects \cite{xia2019extended}.

Based on the success of these recent methods and the simplicity of the PHD filter, the purpose of this paper is to develop a PHD filter for tracking sets of trajectories of extended objects. The objective is to obtain a computationally efficient method that can estimate trajectories of an objects kinematic state as well as its extent, using the entire sequence of measurements up until the current time. The benefit of this is that past estimates of an objects size and shape can be improved. This is achieved by modifying the GGIWPHD filter proposed by Granström et al. \cite{Granstrom2015} to handle sets of trajectories for the kinematic state as well as the extent state. The proposed filter is named the GGIWTPHD filter.


The remaining sections of this paper are structured as follows:
Section \ref{ch:problem_formulation} presents the problem formulation, including the extended MOT problem and all assumptions, Section \ref{ch:EOT_models} presents the models and assumptions made. An overview of the GGIWPHD filter is presented in Section \ref{ch:ggiwphd_filter} as it shares many similarities with the proposed filter. The GGIWTPHD filter equations are presented in Section \ref{ch:ggiwtphd_filter} along with modelling assumptions. Simulated scenarios together with results are presented in Section \ref{ch:simulations} and finally the paper is concluded with a short analysis of the results.

\section{Problem Formulation} \label{ch:problem_formulation}
In short, the objective is to estimate the trajectories of an unknown number of objects (with unknown states) that are in the surveillance area of a sensor. 
Furthermore, the objects of interest are assumed to be extended, i.e., each object can generate multiple measurements from the sensor at each scan. 

More formally, let us assume that there are $N_k$ extended objects present in the surveillance area at time $k$, each described by its extended object state, denoted $\xi_k$. The set of extended objects at time $k$ can, thus, be defined as the RFS,
\begin{equation} \label{eq:multiObjectState}
    \mathbf{X}_k = \left\{ \xi^{(j)}_k \right\}_{j=1}^{N_k},
\end{equation}
where $N_k$ and $\mathbf{X}_k$ are unknown and time-varying. 
Furthermore, let us define a trajectory $\mathcal{T}$ of extended object states for a single object as the tuple: 
\begin{equation}\label{eq:trajectoryState}
    \mathcal{T}_k = (t, \xi^n_{k}) 
\end{equation}
where $\xi^n_{k} = \begin{bmatrix} \xi_t & ... & \xi_k \end{bmatrix}$ is a sequence of states, $t$ denotes the time of birth of the trajectory and $n~=~k~-~t$ denotes the current lifespan of the trajectory. Similarly as in \eqref{eq:multiObjectState}, we can define the set of trajectories present in the surveillance area at time $k$ as the multi-trajectory state
\begin{equation}
    \mathbf{T}_k = \{ \mathcal{T}^{(j)}_k \}_{j=1}^{N_k}.
\end{equation}


If we denote the set of all senor observations up to and including time $k$ as $\mathbf{Z}^k$, the problem considered in this paper is to, based on $\mathbf{Z}^k$, estimate the set of trajectories currently present in the surveillance area. 
In a Bayesian setting, this means that we are interested in recursively calculating the posterior density of the set of trajectories. In theory, we can do this using the standard prediction
\begin{equation}
    p(\mathbf{T}_k|\mathbf{Z}^{k-1}) = \int p(\mathbf{T}_k|\mathbf{T}_{k-1})p(\mathbf{T}_{k-1}|\mathbf{Z}^{k-1})\delta\mathbf{T}_{k-1}
\end{equation}
and update
\begin{align}
    p(\mathbf{T}_k|\mathbf{Z}^k) &= K^{-1} g_k(\mathbf{Z}^k | \mathbf{T}_k)p(\mathbf{T}_k|\mathbf{Z}^{k-1})
    \\
    K &= \int g_k(\mathbf{Z}^{k} | \mathbf{T}_k) p(\mathbf{T}_k|\mathbf{Z}^{k})\delta\mathbf{T}_k
\end{align}
steps, where K is a normalization factor.
Note that all integrals above are set-integrals.

\section{Extended Object Tracking Models} \label{ch:EOT_models}
This section presents the modeling assumptions used in both the GGIWPHD and GGIWTPHD filters. 

\subsection{Extended Object State}
The extended object state $\xi_k$ is denoted as a triple
\begin{equation}
    \xi_k \overset{\Delta}{=} (\gamma_k,\; \mathbf{x}_k,\; X_k)
\end{equation}
where the kinematic state $\mathbf{x}_k = \begin{bmatrix} \mathbf{p}_k & v_k & \phi_k & \omega_k \end{bmatrix}^\mathsf{T} \in \mathbb{R}^5$ describes the 2D-position $\mathbf{p}_k$, speed $v_k$, heading $\phi_k$ and yaw-rate $\omega_k$ of the object, 
the random matrix $X_k \in \mathbb{S}^2_{++}$ as proposed in \cite{Koch2008} models its size and shape by an ellipsoid \cite{Gilholm2005, Granstrom2015}, and 
the measurement rate $\gamma_k > 0$ describes the expected number of measurements generated by the object. 

\subsection{Single object motion models}\label{MotionModel}
This section presents the motion models used in this paper to describe the evolution of the kinematic state, the extension state and the measurement rate. It is assumed that all objects follow the same motion models and that they move independently of each other. 
\subsubsection{Kinematic state}
The motion model for the kinematic state is defined as
\begin{equation}\label{eq:kinematicMotionModel}
    \mathbf{x}_{k+1}=\mathbf{f}\left(\mathbf{x}_{k}\right)+\mathbf{q}_k, \quad \mathbf{q}_{k} \sim \mathcal{N}\left(\mathbf{0}, \mathbf{Q}\right),
\end{equation}
where the state prediction $\mathbf{f}(\mathbf{x}_k)$ and process noise covariance $\mathbf{Q}$ are
\begin{subequations}
    \begin{equation}
    \mathbf{f}\left(\mathbf{x}_{k}\right)= \mathbf{x}_k + \begin{bmatrix} T_s v_k \cos{\phi_k} \\ T_s v_k \sin{\phi_k} \\ 0 \\ T_s \omega_k \\ 0 \end{bmatrix},\\
    \end{equation}
\begin{equation}
\mathbf{Q} = \mathbf{G} \begin{bmatrix} \sigma_v^2 & 0 \\ 0 & \sigma_{\omega}^2 \end{bmatrix} \mathbf{G}^{\top}, \quad
\mathbf{G} = \begin{bmatrix} 0 & 0 & T_s & 0 & 0 \\ 0 & 0 & 0 & 0  & T_s \end{bmatrix}^{\mathsf{T}}.
\end{equation}
\end{subequations}
Here, $T_s$ is the sampling period, $\sigma_v$ is the standard deviation in velocity and $\sigma_{\omega}$ is the standard deviation in yaw-rate. For the filters, we use a linear approximation of \eqref{eq:kinematicMotionModel} where the Jacobian of $\mathbf{f}(\mathbf{x}_k)$ is denoted $F(\mathbf{x}_k)$.

\subsubsection{Extension state}
The random matrix transition density $p(X_{k+1}|\mathbf{x}_k,X_k)$ is conditioned on the kinematic state $\mathbf{x}_k$, which can be utilized to predict rotations of the extension state \cite{Granstrom2015}. The predicted rotation of the extent state is expressed:
\begin{subequations}
\begin{align}
    X_{k+1} &=M\left(\mathbf{x}_{k}\right) X_{k} M\left(\mathbf{x}_{k}\right)^{\top} \\
    M\left(\mathbf{x}_{k}\right) &=\left[\begin{array}{cc}
    \cos \left(\omega_{k} T_{s}\right) & -\sin \left(\omega_{k} T_{s}\right) \\
    \sin \left(\omega_{k} T_{s}\right) & \cos \left(\omega_{k} T_{s}\right)
    \end{array}\right]
\end{align}
\end{subequations}
Using this model, the shape is rotated $\omega_kT_s$ radii to compensate for the objects rotational motion.

The transition density with this model is Wishart
\begin{align}
    p(X_{k+1} &| \mathbf{x}_{k}, X_{k}) = \nonumber\\
    &\mathcal{W}\left(X_{k+1} ; n_e, n_e^{-1} M\left(\mathbf{x}_{k}\right) X_{k} M\left(\mathbf{x}_{k}\right)^{\top}\right),
\end{align}
where $n_e$ denotes the degrees of freedom and represents the uncertainty of the transition where a higher value on $n_e$ means higher certainty \cite{6850178}.

\subsubsection{Measurement rate}
It is assumed that the number of measurements generated from an object stays constant, i.e., 
\begin{equation}
    \gamma_{k+1} = \gamma_k.
\end{equation}
However, to ease this assumption somewhat, the variance of $\gamma_{k+1}$ is increased by a multiplicative factor $\eta_k > 1$ where the closer this factor is to $1$, the more certain the time evolution.




\subsection{RFS processes}\label{sec:RFSProcess}
It is assumed that the multi-object state $\mathbf{X}_{k+1}$ is the union of surviving objects from the previous time instance $\mathbf{X}^s_{k} \subseteq \mathbf{X}_{k}$ and new objects entering the surveillance area $\mathbf{X}^b_{k+1}$. Each object $\xi \in \mathbf{X}_{k}$ is assumed to survive independently of all other objects and with a constant probability of survival $P_S$ and, for the surviving objects, their motion is described using the model detailed in Section \ref{MotionModel}. The arriving objects on the other hand, are modelled by a known multi-object birth process. 
Note that, these survival and birth processes also holds for the time evolution set of trajectories $\mathbf{T}_{k}$.

\subsection{Measurement model}\label{measmodel}
As in \cite{Granstrom2015}, it is assumed that the sensor measures range and bearing to the objects and that they are converted to Cartesian coordinates before being handled by the filters such that a linear measurement model can be used.
The $j^\text{th}$ measurement is, thus, constructed as 
\begin{equation}\label{eq:pol2cart}
    \mathbf{z}_{k}^{(j)}=\left[r_{k}^{(j)} \cos \left(\varphi_{k}^{(j)}\right), \quad r_{k}^{(j)} \sin \left(\varphi_{k}^{(j)}\right)\right]^{\top},
\end{equation}
where $r_{k}^{(j)}$ and $\varphi_{k}^{(j)}$ are the range and bearing measurements corresponding to a position $\mathbf{p}_{k}^{(j)}$. 

Additionally, at each time instance $k$, the sensor returns both a set of object generated measurements $\mathbf{Z}_{k,o}$ as well as a set of clutter detections $\mathbf{Z}_{k,c}$. The total set of measurements $\mathbf{Z}_k$ is the union between these, $\mathbf{Z}_k = \mathbf{Z}_{k,o} \cup \mathbf{Z}_{k,c}$,
where the clutter detections $\mathbf{Z}_{k,c}$ are assumed to be i.i.d over the surveillance area and the number of detections, $|\mathbf{Z}_{k,c}|$, is assumed to be Poisson distributed with a known rate $\lambda_c$. 

For the object generated observations, it is assumed that an object with extended state $\xi_k$ generates $M_k \sim \mathcal{PS}(\gamma_k)$ independent measurements where each measurement is modeled as being uniformly distributed over the surface of the object. The likelihood for a single object-generated measurement $\mathbf{z}_{k}$ can then be expressed as \cite{Granstrom2015}
\begin{align}\label{eq:singleLikelihood}
    p\left(\mathbf{z}_{k} | \xi_{k}\right) =& \; p\left(\mathbf{z}_{k} | \mathbf{x}_{k}, X_{k}\right) \nonumber \\ 
    =& \; \mathcal{N}\left(\mathbf{z}_{k} ; H_{k} \mathbf{x}_{k}, \rho X_{k}+R\left(\mathbf{p}_{k}\right)\right),
\end{align}
where $H_k=\begin{bmatrix} \mathbf{I}_{2\times2} & \mathbf{0}_{2\times3} \end{bmatrix}$ and $\rho$ is a scaling parameter. $R(\mathbf{p})$ is given by a first order Taylor approximation of \eqref{eq:pol2cart} w.r.t. the measurement noise processes 
\begin{subequations}
\begin{align}
    R(\mathbf{p}) &= \mathbf{J}(\mathbf{p}) \operatorname{diag}\left(\left[\sigma_{r}^{2}, \sigma_{\varphi}^{2}\right]\right) \mathbf{J}(\mathbf{p})^{\mathbf{T}} \\
    \mathbf{J}(\mathbf{p}) &= \left[\begin{array}{ll}
    \cos (\varphi) & -r \sin (\varphi) \\
    \sin (\varphi) & r \cos (\varphi)
    \end{array}\right].
\end{align}
\end{subequations}
and models the non-constant across-range noise variance.

Using the assumptions above, the set likelihood of the detections $\mathcal{Z}_{k,o}$ from a single object with state $\xi_k$ is described
\begin{equation}
    P(\mathcal{Z}_{k,o}|\xi_k) = M_k!P(M_k|\xi_k) \prod_{j=1}^{M_k} p(\mathbf{z}_k^{(j)} | \xi_k)
\end{equation}
where $P(M_k|\xi_k) = \mathcal{PS}(M_k;\;\gamma_k)$ and $p(\mathbf{z}_k^{(j)} | \xi_k)$ is in \eqref{eq:singleLikelihood}.

\section{Gamma Gaussian Inverse Wishart PHD Filter}
In this section, we summarize the GGIWPHD filter by Granström et al. \cite{Granstrom2015} to give the basis for understanding the extension to tracking sets of trajectories presented in Section \ref{ch:ggiwtphd_filter}. For the GGIWPHD, the posterior of a single extended object state is assumed to be GGIW distributed with independent components:
\begin{equation} \label{eq:ggiwcomp}
    \begin{aligned}
    p\left(\xi_{k} | \mathbf{Z}^{k}\right) 
    = & \; p\left(\gamma_{k} | \mathbf{Z}^{k}\right) p\left(\mathbf{x}_{k} | \mathbf{Z}^{k}\right) p\left(X_{k} | \mathbf{Z}^{k}\right) \\
    = & \; \mathcal{G}\left(\gamma_{k} ; \alpha_{k | k}, \beta_{k | k}\right) \mathcal{N}\left(\mathbf{x}_{k} ; m_{k | k}, P_{k | k}\right) \\  
     & \; \times \mathcal{IW}_{d}\left(X_{k} ; v_{k | k}, V_{k | k}\right) \\
    = & \; \mathcal{G} \mathcal{G} \mathcal{I} \mathcal{W}\left(\xi_{k} ; \zeta_{k | k}\right)
    \end{aligned}
\end{equation}
where $\zeta_{k | k}=\left\{\alpha_{k | k}, \beta_{k | k}, m_{k | k}, P_{k | k}, v_{k | k}, V_{k | k}\right\}$ is the set of GGIW density parameters.
In this set, $\alpha_{k | k}$ and $\beta_{k | k}$ are the shape and rate parameters of the Gamma distribution, $m_{k | k}$ and $P_{k | k}$ are the mean and variance of the normal distribution, and $v_{k | k}$ and $V_{k | k}$ are the degrees of freedom and scale matrix in the inverse Wishart distribution.

Further, the multi-object state $\mathbf{X}_k$ is assumed to be a Poisson RFS which is described by its PHD intensity. The aim of the GGIWPHD filter is, thus, to approximate its PHD intensity $D_{k | k}$ using the complete measurement sequence $\mathbf{Z}^{k}$ as the GGIW mixture, 
\begin{equation} \label{eq:ggiwmix}
    D_{k | k}\left(\xi_{k}\right)=\sum_{j=1}^{J_{k | k}} w_{k | k}^{(j)} \mathcal{GGIW}\left(\xi_{k} ; \zeta_{k | k}^{(j)}\right)
\end{equation}
where $J_{k|k}$ is the number of components and $w_{k | k}^{(j)}$ is the weight of the $j$:th mixture component.
The parameters of the posterior intensity in \eqref{eq:ggiwmix} are recursively calculated in two steps, prediction and update, as detailed below. 

\subsection{Prediction}
With a prior intensity as \eqref{eq:ggiwmix} and with the modelling assumptions in Section \ref{sec:RFSProcess}, the predicted PHD is also a GGIW mixture consisting of two parts, 
\begin{equation} \label{eq:predPHD}
    D_{k+1 | k}\left(\xi_{k+1}\right)=D_{k+1}^{b}\left(\xi_{k+1}\right)+D_{k+1 | k}^{s}\left(\xi_{k+1}\right)
\end{equation}
where the birth intensity $D_{k}^{b}\left(\xi_{k}\right)$ is a GGIW mixture that models the appearance of new objects in the scene and 
\begin{equation}\label{eq:predSurvivingObjects}
    D_{k+1 | k}^{s}\left(\xi_{k+1}\right) = \sum_{j=1}^{J_{k | k}} w_{k+1 | k}^{(j)} \mathcal{GGIW}\left(\xi_{k+1} ; \zeta_{k+1 | k}^{(j)}\right)
\end{equation}
is the intensity for surviving existing objects. The parameters $\zeta_{k+1 | k}^{(j)}$ for each mixture component in \eqref{eq:predSurvivingObjects} are obtained from the prior using the dynamic motion models detailed in Section \ref{MotionModel} and its weight is updated according to the probability of survival, $w_{k+1|k}^{(j)} = P_{S} w_{k | k}^{(j)}$. For more details regarding the prediction step of the GGIWPHD filter, see \cite{Granstrom2012}.

\subsection{Update}
The predicted PHD intensity in \eqref{eq:predPHD} is updated using the current observations $\mathbf{Z}_k$ to form the posterior PHD intensity consisting of three components 
\begin{equation} \label{eq:updPHD}
    D_{k | k}\left(\xi_{k}\right)=D_{k | k}^{m}\left(\xi_{k}\right)+D_{k | k}^{b}\left(\xi_{k}\right)+D_{k | k}^{d}\left(\xi_{k}\right)
\end{equation}
where $D_{k | k}^{m}$ represents undetected previously existing objects, $D_{k | k}^{b}$ the new objects and $D_{k | k}^{d}$ the detected previously existing objects. 
In the following part of this section, the three parts of the posterior PHD will be explained further.

\subsubsection{Undetected objects}
The PHD for undetected previously existing objects is simply the predicted intensity in \eqref{eq:predSurvivingObjects} but where the mixture weights and the measurement rate parameters are adjusted due to the missed detections. The updated intensity for undetected objects then becomes
\begin{equation}
    D^m_{k|k} \approx \sum^{J_{k|k}}_{j=1} \tilde{w}_{k|k}\mathcal{GGIW}(\xi_k;\tilde{\zeta}^{(j)}_{k|k})
\end{equation}
where $\tilde{\zeta}^{(j)}_{k|k} = \{ \tilde{\alpha}^{(j)}_{k|k},~ \tilde{\beta}^{(j)}_{k|k},~ m^{(j)}_{k|k-1},~ P^{(j)}_{k|k-1},~ v^{(j)}_{k|k-1},~ V^{(j)}_{k|k-1} \}$
and $\tilde{\alpha},~ \tilde{\beta},~ \tilde{w}$ are obtained using gamma-mixture reduction \cite{6290567}. For more details regarding calculations of these parameters, refer to \cite{Granstrom2015}.

\subsubsection{Detected objects}
To handle the data association uncertainty for extended objects, the set $\mathbf{Z}_k$ is clustered into a number of cells where each cell $\mathbf{W}$ represents possible measurements from one extended object. One such clustering is called a partition $\mathcal{P}\angle\mathbf{Z}_k$. For the measurement update to be tractable, only a subset of all partitions can be considered \cite{Granstrom2012}. It is common to use spatial clustering algorithms to group measurements into partitions to retrieve this subset.

Using these partitionings, 
new objects are born at the center of each cell (i.e., a measurement cluster) and are represented by the following PHD
\begin{equation} \label{eq:newobj}
    D_{k | k}^{b}\left(\xi_{k}\right)= \sum_{\mathcal{P} \angle \mathbf{Z}_{k}} \sum_{\mathbf{W} \in \mathcal{P}} w_{k | k}^{(b, \mathbf{W})} \mathcal{GGIW}\left(\xi_{k} ; \zeta_{k | k}^{(b, \mathbf{W})}\right)
\end{equation}
where the density parameters $\zeta_{k | k}^{(b, \mathbf{W})}$ are updated using predefined birth parameters 
\begin{equation}\zeta_{k|k}^{(b)}=\left\{\alpha_{k|k}^{(b)}, \beta_{k|k}^{(b)}, m_{k|k}^{(b)}, P_{k|k}^{(b)}, v_{k|k}^{(b)}, V_{k|k}^{(b)}\right\}. 
\end{equation}

Similarly, 
the detected previously existing objects are updated with the same partitionings and the measurement model in Section \ref{measmodel} to be represented by the following PHD
\begin{equation} \label{eq:prevobj}
    D_{k | k}^{d}\left(\xi_{k}\right)=\sum_{\mathcal{P} \angle \mathbf{Z}_{k}} \sum_{\mathbf{W} \in \mathcal{P}} \sum_{j=1}^{J_{k | k-1}} w_{k | k}^{(j, \mathbf{W})} \mathcal{G} \mathcal{G} \mathcal{I} \mathcal{W}\left(\xi_{k} ; \zeta_{k | k}^{(j, \mathbf{W})}\right)
\end{equation}
A more comprehensive explanation of the measurement update is given in \cite{Granstrom2015}. 

\subsection{Post-processing}\label{sec:ggiwphd_post_processing}
Equations \eqref{eq:predPHD} and \eqref{eq:updPHD} shows the prediction and update causes the number of hypotheses to grow exponentially. Therefore methods to reduce the number of hypothesis are used. In the GGIWPHD filter pruning, merging and capping is used to for hypothesis reduction.

Additionally, to extract estimates of likely extended object states, the set of GGIW components is extracted for which $w_{k|k}^{(j)}>\Bar{w}_e$ where $\Bar{w}_e$ is a threshold. In short, this step provides 
\begin{equation} \label{eq:ggiw_extract}
    \hat{\mathbf{X}}_{k | k}=\left\{\hat{\xi}_{k | k}^{(i)}\right\}_{i: 1}^{\hat{N}_{k | k}}, \quad \hat{\xi}_{k | k}^{(i)}=\left(\mathbb{E}\left[\gamma_{k}\right], \mathbb{E}\left[\mathbf{x}_{k}\right], \mathbb{E}\left[X_{k}\right]\right)
\end{equation} 
Note that, it is not clear which of these estimates correspond to which objects over time. 
\label{ch:ggiwphd_filter}

\section{Tracking sets of trajectories}\label{ch:SetOfTrajectories}
As with most PHD filters, the PHD intensity in the GGIWPHD filter does not put labels on the objects in the surveillance area and, as a consequence, there is no formal process for building object trajectories. Instead, ad-hoc methods to label the GGIW components have been proposed, e.g. \cite{Granstrom2015}, which 
are susceptible to track switches, missed detections and false tracks \cite{Garcia-Fernandez2018}. In this section, we present the additional assumptions needed to adapt the GGIWPHD filter for extended objects to directly estimate the sets of trajectories as been successfully done for point-source targets PHD \cite{Garcia-Fernandez2018} and the extended object PMBM filter \cite{xia2019extended}.

Similarly as in the GGIWPHD, we assume that the posterior for a single trajectory can be factorized into three independent densities, 
\begin{equation} \label{ggiwtdensity}
\begin{aligned}
    p(\mathcal{T}_k | \mathbf{Z}^k) = & \; p(\gamma_k | \mathbf{Z}^k) p(\mathbf{x}^n_{k}|\mathbf{Z}^k) p(X^n_{k}| \mathbf{Z}^k) \\
    = & \; \mathcal{G}\left(\gamma_{k} ; \alpha_{k | k}, \beta_{k | k}\right)\mathcal{N}\left(\mathbf{x}^n_{k| k} ; m^n_{k | k}, P^n_{k | k}\right) \\  
      & \; \times \mathcal{IW}_{d}\left(X^n_{k | k} ; v^n_{k | k}, V^n_{k | k}\right)\\
    = \;& \mathcal{GGIWT}\left(\mathcal{T}_k ; \zeta_{k | k}\right).
\end{aligned}
\end{equation}
where $\zeta_{k|k}$ represents the GGIWT density parameters
\begin{equation}
    \zeta_{k|k}=\left\{\alpha_{k | k}, \beta_{k | k}, m^n_{k | k},  P^n_{k | k}, v^n_{k | k}, V^n_{k | k}\right\}.
\end{equation}
Note that, compared to the GGIW distribution, the GGIWT components describe a distribution over a trajectory of $n$ kinematic state parameters ($m^n_{k | k}$, $P^n_{k | k}$) and extended state parameters ($v^n_{k | k}$, $V^n_{k | k}$), where as the measurement rate parameters ($\alpha_{k|k}$, $\beta_{k|k}$) are assumed constant and, thus, not treated as trajectories.


Furthermore, the multi-trajectory state $\mathbf{T}_k$ is assumed to be a Poisson RFS. This implies that the number of trajectories, cardinality of $\mathbf{T}_k$, is Poisson distributed and that, for each cardinality, the trajectories are IID. The multi-trajectory density thus has the form 
\begin{equation}\label{eq:multiTrajDensity}
    \pi(\mathbf{T}_k | \mathbf{Z}^k) = e^{-\lambda}\lambda^{N_k}\prod^{N_k}_{j=1}\mathcal{GGIWT}\left(\mathcal{T}_k^{(j)}; \zeta^{(j)}_{k | k} \right)
\end{equation}
where $\lambda^{N_k} \geq 0$ is the expected number of trajectories. As with all Poisson RFSs, the multi-state density in \eqref{eq:multiTrajDensity} is characterised by its PHD $D_{k|k}(\mathcal{T}_k)$ for which we propose the method GGIWTPHD filter in Section \ref{ch:ggiwtphd_filter}.

\section{Gamma Gaussian Inverse Wishart Trajectory PHD Filter}
\label{ch:ggiwtphd_filter}

For the assumptions outlined in Sections \ref{ch:EOT_models} and \ref{ch:SetOfTrajectories}, the aim of the GGIWTPHD filter is to recursively and at each time instance describe the multi-trajectory density in \eqref{eq:multiTrajDensity} by approximating its PHD on the following form,
\begin{equation}
    D_{k|k} = \sum_{j=1}^{J_k}w^{(j)}_{k|k}\mathcal{GGIWT}\left( \mathcal{T}^{(j)}_k; \zeta_{k|k}^{(j)}\right)
\end{equation}
Compared to the GGIWPHD filter, the GGIWTPHD filter estimates the trajectory over the kinematic and extension states while the same assumptions regarding the process and measurements, made in Section \ref{sec:RFSProcess}, still hold. The GGIWTPHD has an added step where smoothing of the extension trajectory is applied after estimation to achieve improved extent estimates over time. The filter recursion of the GGIWTPHD filter is shown in Fig. \ref{fig:GGIWTPHD-flowchart}. However, the prediction and update steps in the GGIWTPHD filter share many similarities to the GGIWPHD filter. This section highlights the differences between the two. For further details of regarding the full set of equations of the GGIWTPHD filter, see \cite{GGIWT_techreport}.


\tikzstyle{startstop} = [rectangle, minimum width=2cm, minimum height=0.8cm,text centered, draw=black]
\tikzstyle{arrow} = [thick,->,>=stealth]
\usetikzlibrary{calc}

\begin{figure}
    \centering
    \scalebox{0.8}{%
     \begin{tikzpicture}[node distance = 0.8cm, auto]
        \node (predict) [startstop] {Prediction};
        \node (update) [startstop, above of=predict, yshift=0.6cm] {Update};
        \node (reduction) [startstop, right of=update, xshift=2cm, yshift=-0.7cm] {Reduction};
        \node (estimation) [startstop, right of=reduction, xshift=1.5cm] 
            {\begin{tabular}{l}
                Trajectory\\
                extraction
            \end{tabular}};
        
        \node (smoothing) [startstop, right of=estimation, xshift=2.2cm] {\begin{tabular}{l}
            Extension\\
            smoothing
        \end{tabular}};
        \node (end) [startstop, below of=smoothing, yshift=-0.4cm, draw=none, minimum size=0cm] {};
        
        \draw [arrow] (predict) -- node {$D_{k|k-1}$} (update);
        \draw [arrow] (update) -- node {$D_{k|k}$} (reduction);
        \draw [arrow] (reduction) -- (estimation);
        \draw [arrow] (reduction) -- node {$k=k+1$} (predict);
        \draw [arrow] (estimation) -- node {$\mathbf{\hat{T}}_{k|k}$} (smoothing);
        \draw [arrow] (smoothing) -- node {$\mathbf{\breve{T}}_{k|k}$} (end);
    \end{tikzpicture}
    }
    \caption{Flowchart of the GGIWTPHD filter recursion. }
    \label{fig:GGIWTPHD-flowchart} 
\end{figure}
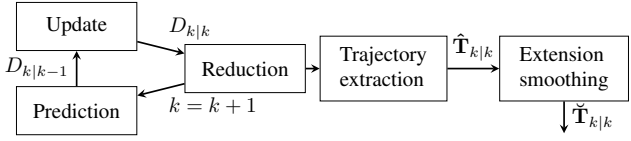

How the different steps in the GGIWTPHD recursion outlined in Fig. \ref{fig:GGIWTPHD-flowchart} differ from the GGIWPHD filter is presented in the following sections. 




\subsection{Prediction}
Due to the shared motion model presented in Section \ref{MotionModel}, the predicted GGIWT mixture is identical in composition to the GGIW predicted mixture in that it is the sum of newly arriving objects and objects that survive from the previous time instance, 
\begin{equation}\label{eq:predGGIWTPHD}
    D_{k+1 | k}\left(\mathcal{T}_{k+1}\right)=D_{k+1}^{b}\left(\mathcal{T}_{k+1}\right)+D_{k+1 | k}^{s}\left(\mathcal{T}_{k+1}\right)
\end{equation}
In this case, the newly arriving objects will initiate new (single state) trajectories at the current time which is described by the trajectory birth PHD $D_{k+1}^{b}\left(\cdot\right)$. As such, the trajectory birth PHD is assumed to be a GGIWT-mixture of (single state) trajectories whose parameters need to be chosen for the specific problem. 
The surviving object PHD $D_{k+1 | k}^{s}\left(\cdot\right)$ is a GGIWT-mixture whose measurement rate parameters are calculated identical as in the GGIWPHD filter while the kinematic and extended state parameters are predicted as%
\footnote{Note that the conditional dependency on the previous time step $k$ is omitted for brevity}
\begin{subequations}
\begin{align}
    m^{n,(j)}_{k+1} &= \begin{bmatrix} (m^{n-1,(j)}_{k})^\mathsf{T}, & (\Dot{F}_{k+1}^{(j)}m^{n-1,(j)}_{k})^\mathsf{T} \end{bmatrix}^\mathsf{T}
    \\
    P_{k+1}^{n,(j)} &= \begin{bmatrix} P_{k}^{n-1,(j)} & P_{k}^{n-1,(j)}(\Dot{F}_{k+1}^{(j)})^\mathsf{T} \\ \Dot{F}_{k+1}^{(j)}P^{n-1,(j)}_{k} & \Dot{F}_{k+1}^{(j)}P_{k}^{n-1,(j)}(\Dot{F}_{k+1}^{(j)})^\mathsf{T} + Q \end{bmatrix}
    \\
    v^{n,(j)}_{k+1} &= \begin{bmatrix} v^{n-1,(j)}_{k}, & v^{(j)}_{k} \end{bmatrix}\label{eq:pred_v}
    \\
    V^{n,(j)}_{k+1} &= \begin{bmatrix} V^{n-1,(j)}_{k}, & V^{(j)}_{k} \end{bmatrix}\label{eq:pred_V}
    \\
    \Dot{F}_{k+1}^{(j)} &= \begin{bmatrix} 0_{1, n-1}, & 1 \end{bmatrix} \otimes F^{(j)}( \tau( m^{n-1,(j)}_{k} ) )
\end{align}
\end{subequations}
where $\otimes$ refers to the Kronecker product and $0_{m,n}$ is a $m \times n$ zero matrix. $\tau(m_{k}^{n})$ is a function that extracts the last state in the trajectory, see \cite{GGIWT_techreport}. 
Further, the computation of $v^{(j)}_{k}$ and $V^{(j)}_{k}$ in \eqref{eq:pred_v} and \eqref{eq:pred_V}, respectively, are identical to the GGIWPHD filter and are left out for brevity.

\subsection{Update}
As the prediction step, the updated GGIWT-mixture is identical in structure to that of the GGIWPHD filter,
\begin{equation}\label{eq:updateGGIWTPHD}
    D_{k | k}\left(\mathcal{T}_{k}\right)=D_{k | k}^{m}\left(\mathcal{T}_{k}\right)+D_{k | k}^{b}\left(\mathcal{T}_{k}\right)+D_{k | k}^{d}\left(\mathcal{T}_{k}\right)
\end{equation}
with undetected objects $D_{k | k}^{m}(\cdot)$, detected newly formed trajectories $D_{k | k}^{b}(\cdot)$, and detected existing trajectories $D_{k | k}^{d}(\cdot)$. 
Again, due to the shared assumptions, many of the equations in the GGIWTPHD filter update are identical to the GGIWPHD filter.
Since the update for undetected objects only updates the parameters related to the measurement rate and the weight of each mixture component, which are not treated as trajectories, this update is the same as in the GGIWPHD filter. For the update of persistent trajectories using cell $\mathbf{W}$ the density parameters are

\begin{subequations}\label{eq:upd_traj}
\begin{align}
    m^{n,(j,\mathbf{W})}_{k|k} &= m^{n,(j)}_{k|k-1} + K_{k | k-1}^{(j, \mathbf{W})} (\bar{\mathbf{z}}_{k}^{\mathbf{W}} - \dot{H}^{(j)}m_{k|k-1}^{n,(j)}) \label{eq:upd_m}
    \\
    P^{n,(j,\mathbf{W})}_{k|k} &= P_{k|k-1}^{n,(j)} - K_{k | k-1}^{(j, \mathbf{W})}\dot{H}P_{k|k-1}^{n,(j)} \label{eq:upd_P}
    \\
    v_{k | k}^{n,(j, \mathbf{W})} &= \begin{bmatrix} v_{k|k-1}^{n-1,(j)}, & v_{k | k}^{(j, \mathbf{W})} \end{bmatrix} \label{eq:upd_v}
    \\ 
    V_{k | k}^{n,(j, \mathbf{W})} &= \begin{bmatrix} V_{k|k-1}^{n-1,(j)}, & V_{k | k}^{(j, \mathbf{W})}\end{bmatrix} \label{eq:upd_V}
    \\
    S_{k|k-1}^{(j,\mathbf{W})} &= \dot{H}P_{k|k-1}^{n,(j)}\dot{H}^\mathsf{T} + \frac{\hat{R}_{k|k-1}^{(j,\mathbf{W})}}{|\mathbf{W}|}
    \\
    K_{k | k-1}^{(j, \mathbf{W})} &= P_{k | k-1}^{n,(j)} \Dot{H}_{k}^{\mathsf{T}}\left(S_{k | k-1}^{(j, \mathbf{W})}\right)^{-1}
    \\
    \hat{R}_{k|k-1}^{(j,\mathbf{W})} &= \rho \hat{X}_{k|k-1}^{(j)} + R\left(\dot{H}m^{n,(j)}_{k|k-1}\right)
    \\
    \hat{X}_{k|k-1}^{(j)} &= V_{k|k-1}^{(j)}\left(v_{k|k-1}^{(j)}-2 d-2\right)^{-1}
    \\
    \dot{H} &= \begin{bmatrix} 0_{1,n-1}, & 1 \end{bmatrix} \otimes H
\end{align}
\end{subequations}
where $\bar{\mathbf{z}}_k^{\mathbf{W}}$ is the centroid measurement of a cell $\mathbf{W}$. The computation of $v_{k | k}^{(j, \mathbf{W})}$ and $V_{k | k}^{(j, \mathbf{W})}$ in \eqref{eq:upd_v} and \eqref{eq:upd_V}, respectively, are identical to the GGIWPHD filter and are left out for brevity. Details about these and more regarding the update equations in the GGIWTPHD filter see \cite{GGIWT_techreport}. From \eqref{eq:upd_m} and \eqref{eq:upd_P} it can be seen that smoothing of the kinematic estimates is applied in the update while the updated parameters for the extension are only concatenated with previous time-steps. To improve the estimate of the extent trajectory, extension smoothing is applied after extraction as outlined in Fig. \ref{fig:GGIWTPHD-flowchart}. 

\subsection{Post-processing}\label{sec:ggiwt_post_processing}
Similar to the GGIW mixture, the GGIWT mixture needs to be managed in order to prevent the filter from becoming computationally intractable. Therefore, capping and pruning is used as in the GGIWPHD filter. For the GGIWTPHD filter, absorption is used instead of merging, for a discussion on why this is advisable and implementation details, refer to \cite{Garcia-Fernandez2018} and \cite{GGIWT_techreport}. Trajectories are extracted from the reduced GGIWT mixture in the same manner as in \eqref{eq:ggiw_extract} and are denoted $\mathbf{\hat{T}}_{k|k}$. Smoothing is then performed on the extent estimates in $\mathbf{\hat{T}}_{k|k}$ and the set of smoothed estimated trajectories $\mathbf{\breve{T}}_{k|k}$ that contain both the kinematic- and extent estimates are returned. When smoothing the extent estimates the method proposed by Granström and Bramstång is used \cite{Granstrom2019}. This requires that the predicted parameters of the mixture at each time step is stored as well as the updated parameters. The proposed smoothing procedure is consistent in a Bayesian setting and the smoothing equations are presented in \cite{GGIWT_techreport}. 

\section{Evaluation} \label{ch:simulations}
\begin{figure}
    \centering
    \includegraphics[trim=10 30 50 66, clip, width=1\linewidth]{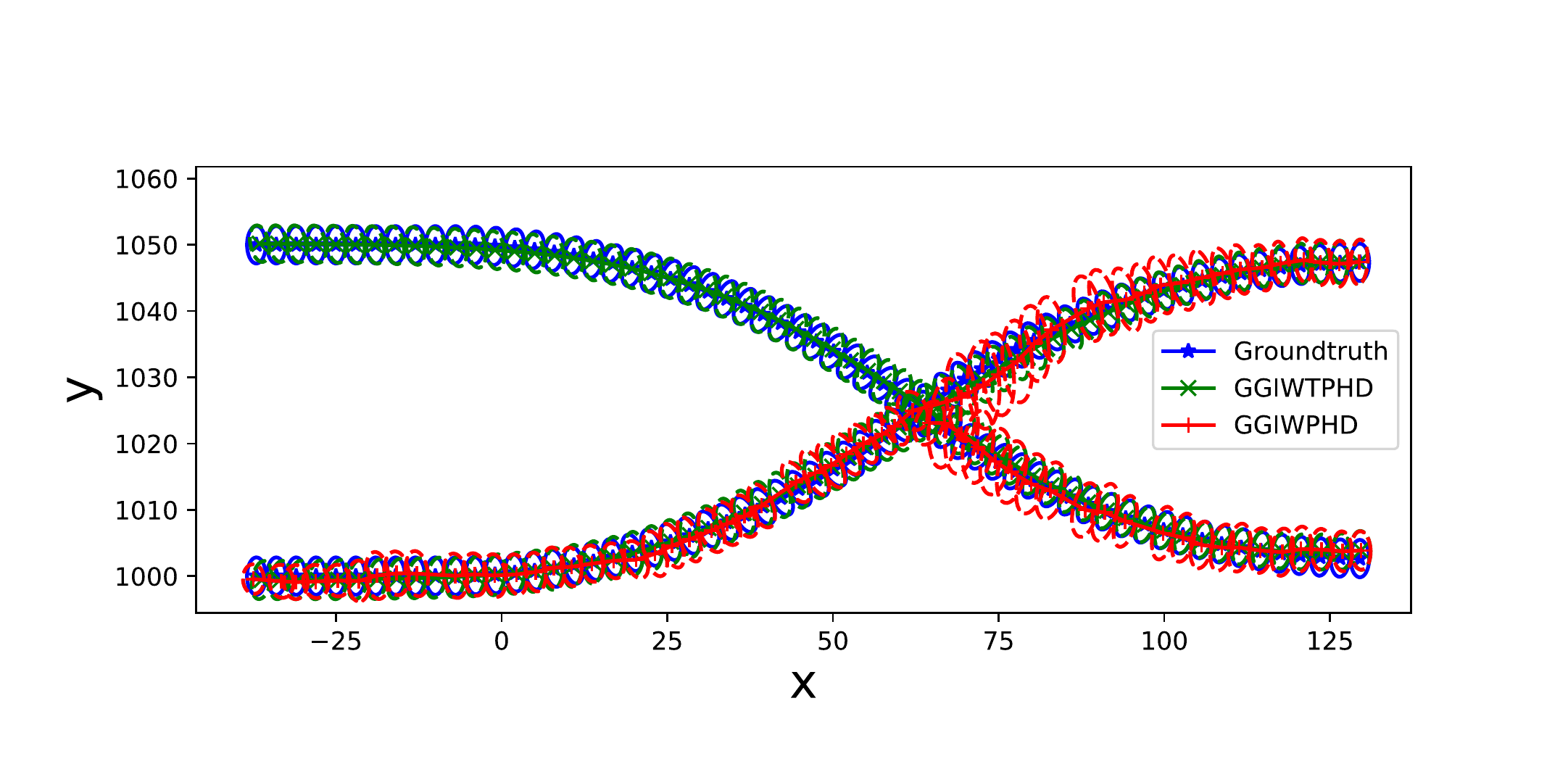}
    \caption{Scenario 1. Two targets are born at time $k=0$ and move in parallel until turning towards each other such that they cross paths after some time. Both die at time $k=60$. Depicted are also estimates from the GGIWPHD filter in red and GGIWTPHD filter in green.}
    \label{fig:scenario_4_estimates}
\end{figure}

\begin{figure}
    \centering
    \includegraphics[trim=0 0 0 0, clip, width=0.68\linewidth]{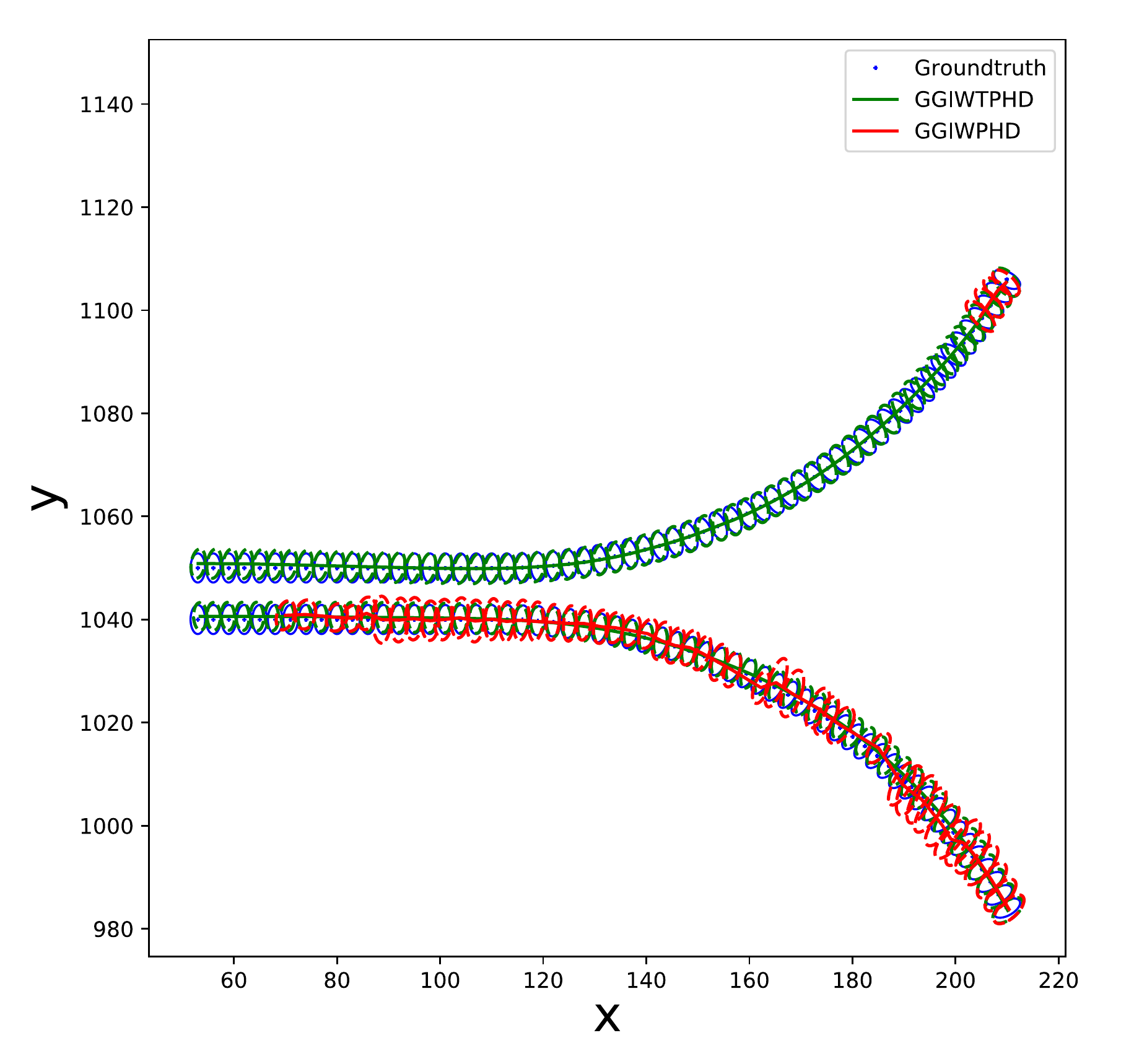}
    \caption{Scenario 2. Two targets are born close together at $k=0$ that move in parallel. At time $k=20$ the objects move apart and die at $k=60$. The figure showcases one scenario where the GGIWPHD filter fails to estimate trajectories while the GGIWTPHD filter succeeds.}
    \label{fig:scenario_3_estimates}
\end{figure}

We compare the performance of the proposed GGIWTPHD filter, with and without smoothing of the extent state, with the GGIWPHD filter. The comparison is made on simulated data from the two challenging scenarios depicted in Fig. \ref{fig:scenario_4_estimates} and Fig. \ref{fig:scenario_3_estimates}. Results from two more scenarios are presented in \cite{GGIWT_techreport} but are omitted here due to page constraints.
The ellipses' major axis is assumed to be aligned with its velocity vector. Each time-step, a Poisson distribution is sampled using a ground truth measurement rate as to obtain the number of measurements a given object should generate. This sample is then used to decide how many normally distributed measurements around the position of the object should be generated. The parameters used for the simulations and filters are summarized in Table \ref{tab:parametersettings}.

\begin{table}
    \caption{Parameter settings for test scenarios.}
    \label{tab:parametersettings}
    \centering
    \resizebox{0.7\columnwidth}{!}{
    \begin{tabular}{|l c|l|} 
        \hline
        \multicolumn{2}{|l|}{\textbf{Parameter}} & \textbf{Value} \\ \hline
        Sampling time                & $T_s$                          & $1$  \\
        Forgetting factor            & $\eta_k$                       & $2$ \\
        Kinematics noise             & $\sigma_a,\sigma_{\omega}$     & $0.2$, $0.2\frac{\pi}{180}$ \\
        Measurement noise            & $\sigma_r,\sigma_{\varphi}$    & $1$, $0.01\frac{\pi}{180}$ \\
        Extension uncertainty        & $n_e$                          & $120$ \\
        Scaling parameter            & $\rho$                         & $0.75$ \\
        Clutter rate                 & $\lambda_c$                    & $100$ \\
        Detection probability        & $P_D$                          & $0.99$ \\
        Survival probability         & $P_S$                          & $0.99$ \\
        Birth weight                 & $w_k^{(b)}$                    & $0.03$ \\   
        Extraction threshold         & $\Bar{w}_e$                    & $0.5$ \\
        Pruning threshold            & $T$                            & $0.001$ \\
        Merging/absorption threshold & $U$                            & $5$ \\
        Capping threshold            & $M$                            & $50$ \\
        Clustering algorithm         &                                & DBSCAN \cite{10.5555/3001460.3001507} \\
        \hline 
    \end{tabular}
    }
\end{table}

\begin{figure}
    \centering
    \subfloat[RMS error of the trajectory metric\label{fig:scenario_4_dospa3d}]{
        \includegraphics[trim=0 0 30 40, clip, height=3cm]{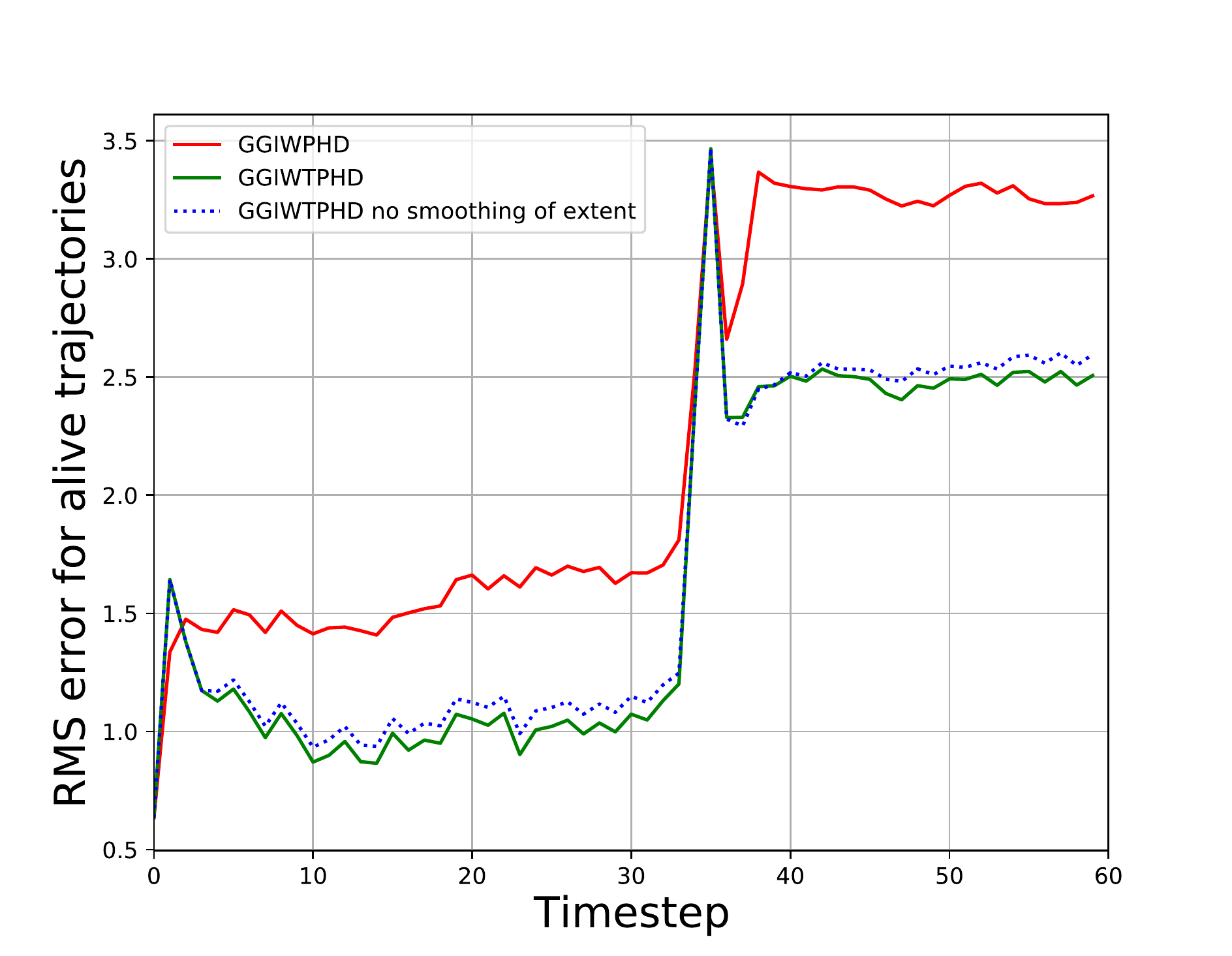}
    }
    \subfloat[Average cardinality\label{fig:scenario_4_avgcard}]{
        \includegraphics[trim=0 0 30 40, clip, height=3cm]{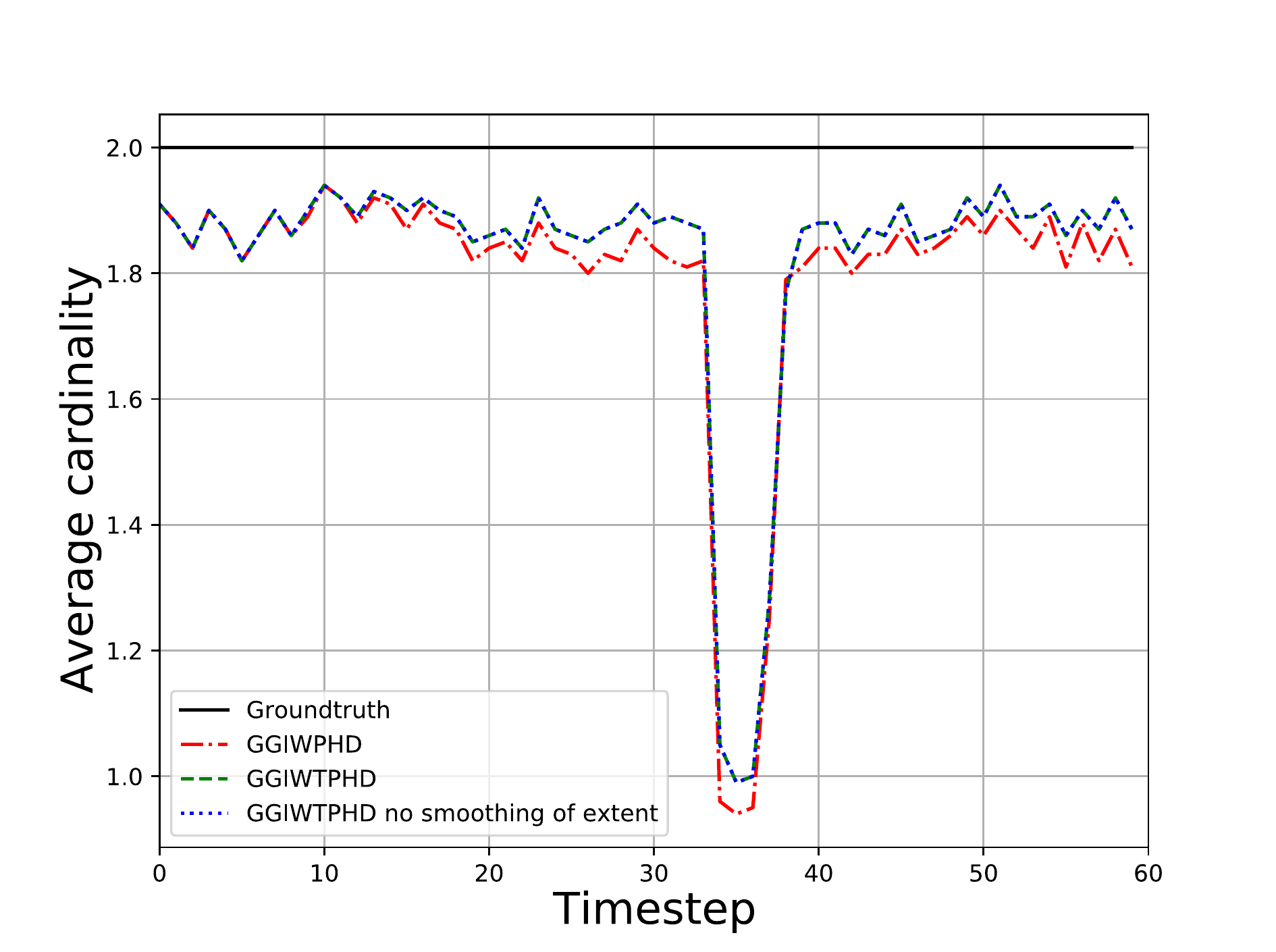}
    }
    \newline
    \subfloat[Decomposed RMS costs\label{fig:scenario_4_decompcost}]{
        \includegraphics[trim=10 60 70 115, clip, width=0.9\linewidth]{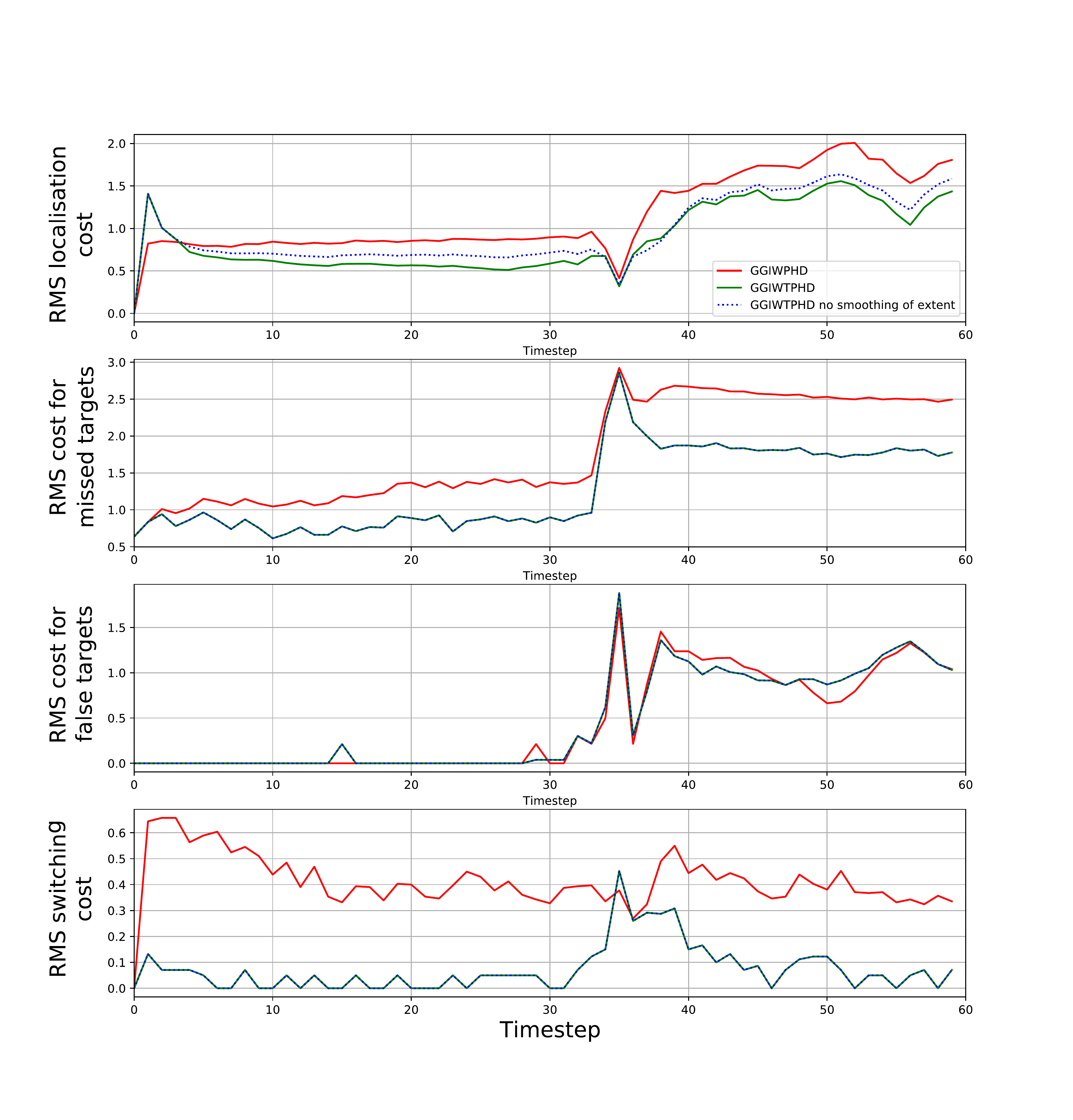}
    }
    \caption{Results comparing GGIWTPHD with smoothing of extent estimates (green), GGIWTPHD without smoothing of extent estimates (dotted blue) and GGIWPHD (red) using labeled components. These are results from a Monte Carlo simulation in scenario 1 of 100 runs.}
    \label{fig:scenario_4_trajmetricMC}
\end{figure}

\begin{figure}
    \centering
    \subfloat[RMS error of the trajectory metric\label{fig:scenario_3_dospa3d}]{
        \includegraphics[trim=10 0 10 40, clip, height=3cm]{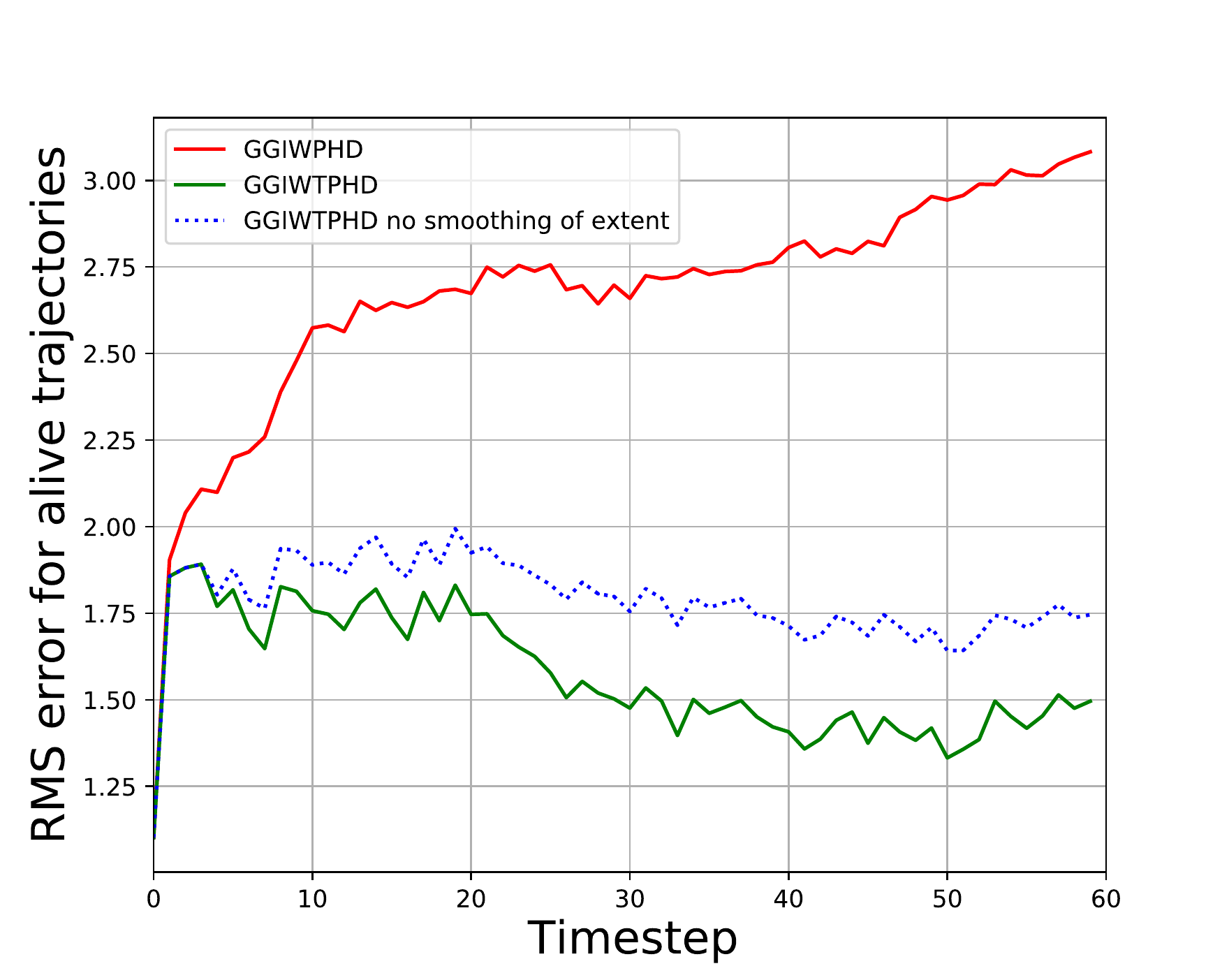}
    }
    \subfloat[Average cardinality\label{fig:scenario_3_avgcard}]{
        \includegraphics[trim=10 0 10 40, clip, height=3cm]{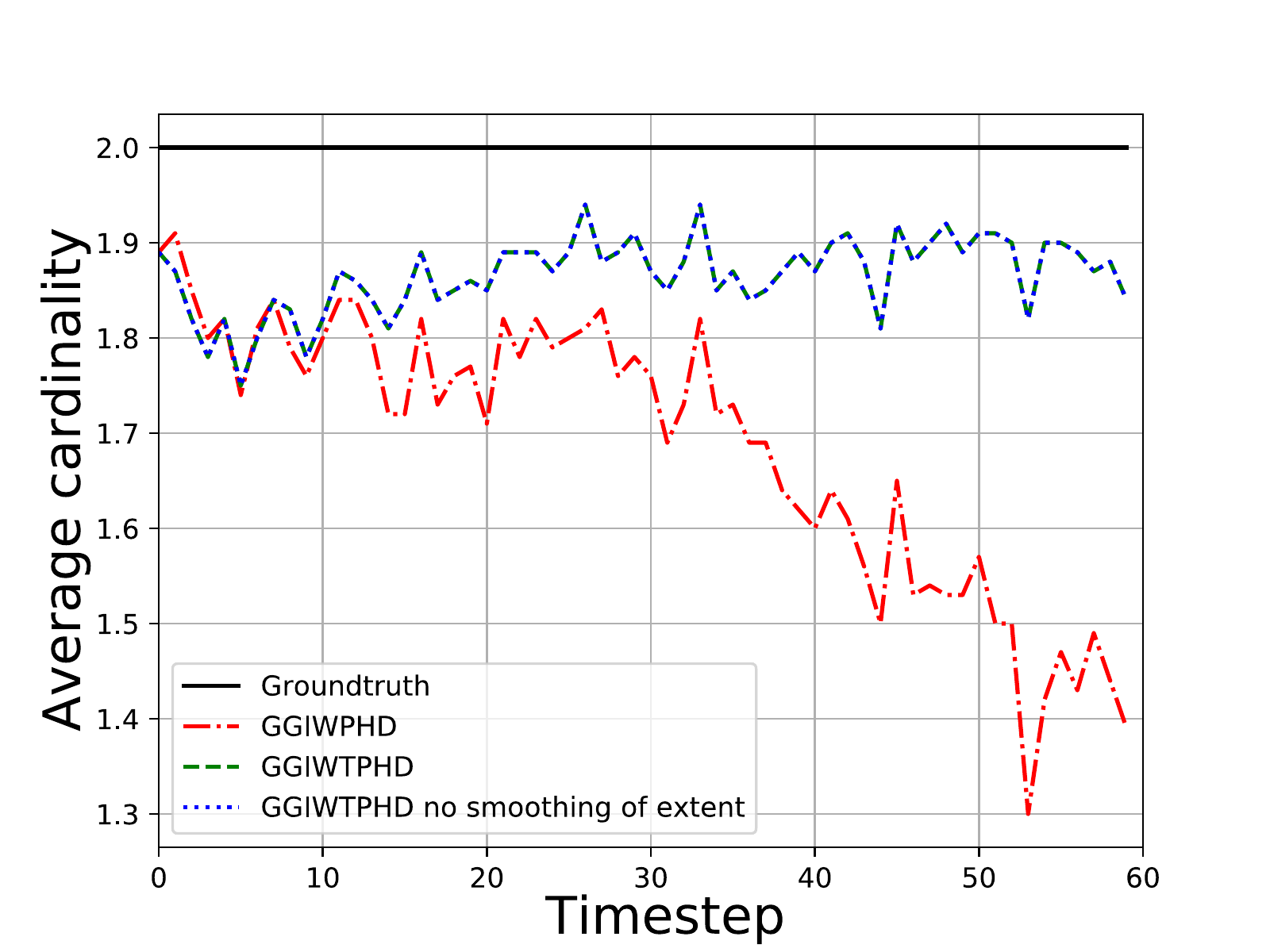}
    }
    \newline
    \subfloat[Decomposed RMS costs\label{fig:scenario_3_decompcost}]{
        \includegraphics[trim=10 10 70 115, clip, width=0.9\linewidth]{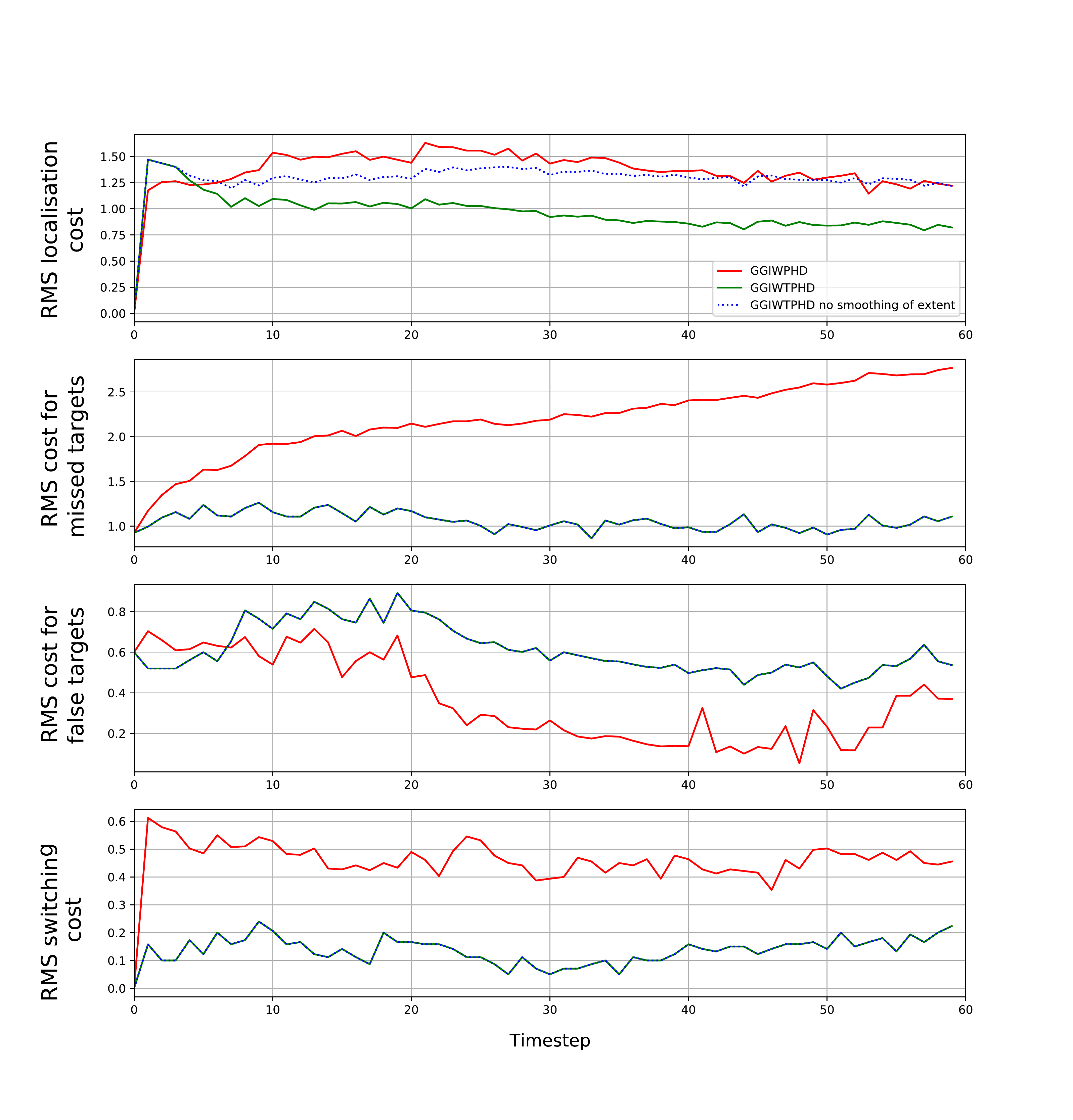}
    }
    \caption{Results comparing GGIWTPHD with smoothing of extent estimates (green), GGIWTPHD without smoothing of extent estimates (dotted blue) and GGIWPHD using labeled components (red). These are results from a Monte Carlo simulation in scenario 2 of 100 runs.}
    \label{fig:scenario_3_trajmetricMC}
\end{figure}

To evaluate the results, a modified metric that combines the GOSPA metric for trajectories presented by Rahmathullah et. al. \cite{9127194} and the Gaussian-Wasserstein Distance \cite{5977698} is used. The combined metric is constructed as, 
\begin{equation}
\begin{aligned}
    &d^2(\mathbf{T}_k, \mathbf{\hat{T}}_k) =\\ &c_l^2(\mathbf{T}_k, \mathbf{\hat{T}}_k) + c_m^2(\mathbf{T}_k, \mathbf{\hat{T}}_k) + \\&c_f^2(\mathbf{T}_k, \mathbf{\hat{T}}_k) + c_t^2(\mathbf{T}_k, \mathbf{\hat{T}}_k),
\end{aligned}
\end{equation}
where $c_l$ is the Gaussian-Wasserstein distance and $c_m$, $c_f$ and $c_t$ are the missed, false and switch costs respectively. Using this decomposition allows for a more comprehensive overview of the performance of the different filters and evaluates how well the filters estimates the cardinality and kinematic state (GOSPA) as well as the object extent (Gaussian-Wasserstein).
The results in term of RMS distance of 100 Monte Carlo simulations on scenario 1 and scenario 2 are shown in Fig. \ref{fig:scenario_4_trajmetricMC} and Fig. \ref{fig:scenario_3_trajmetricMC}, respectively. 

The defined metric shows that in scenario 1, both filters have issues estimating the trajectories of the objects at time $k\approx35$ when the objects cross. This is likely due to the wrong measurement clusters being computed for that timestep which will affect the filters' performances. It can also be seen from the decomposed metric that the GGIWTPHD filter suffers less from track switching before and after time $k\approx35$ which indicates that the filter is more capable of keeping single trajectories alive without switching.

In scenario 2 it is likely that during the first part of the sequence ($k \leq 20$) the measurements from the two objects will be clustered together as one for many of the time-steps. This can be seen from the fact that the GGIWPHD filter seems to underestimate cardinality while the GGIWTPHD filter does not to the same extent. This is likely the reason why the GGIWPHD filter gives a lower cost for false targets but a higher cost for missed targets in the metric.

Overall the total RMS error for the GGIWTPHD filter is lower for both scenario 1 and 2. In scenario 1 they become similar at time $k\approx35$ but then the GGIWTPHD filters RMS drops down below the GGIWPHD filter which indicates that it is capable of recovering from difficult situations better than the GGIWPHD filter.

\section{Conclusion}
This paper has presented a new algorithm for tracking sets of trajectories of extended objects which is named the GGIWTPHD filter. The filter is derived using a heuristic approach that combines the results presented in three different papers. The first concerns the construction of trajectories for the kinematic state \cite{Garcia-Fernandez2018}. The second implements a PHD filter using a Gamma Gaussian inverse Wishart PHD \cite{Granstrom2015} and the third concerns Bayesian smoothing of extent estimates in the random matrix framework \cite{Granstrom2019}. The filter recursions estimates the set of alive trajectories at each time step and then performs smoothing on the set of extent estimates at each time step.

Monte Carlo evaluation on two challenging scenarios using a new proposed metric shows that the GGIWTPHD filter with smoothing performs better in both scenarios. All evaluated algorithms suffer a decrease in performance when incorrect clusters are computed but it is also shown that even in this case, the GGIWTPHD filter outperforms the GGIWPHD filter.

\FloatBarrier
\printbibliography

\newpage

\end{document}